\pdfoutput=1
\documentclass[11pt]{article}

\usepackage[review]{acl}

\usepackage{times}
\usepackage{latexsym}

\usepackage{graphicx} 
\usepackage{float} 
\usepackage{subfigure}

\usepackage[T1]{fontenc}
\usepackage[utf8]{inputenc}

\usepackage{microtype}

\title{Analyzing Folktales of Different Regions Using Topic Modeling and Clustering }

\author{Jacob Werzinsky \qquad Zhiyan Zhong \qquad Xuedan Zou \\
  Department of Computer Science \\
  Dartmouth College \\
  \texttt{\{jacob.e.werzinsky.22,zhiyan.zhong.gr,xuedan.zou.gr\}@dartmouth.edu} \\
}

\begin{document}
\nolinenumbers
\maketitle
\begin{abstract}

%Basic intro:
This paper employs two major natural language processing techniques, topic modeling and clustering, to find patterns in folktales and reveal cultural relationships between regions. \ In particular, we used Latent Dirichlet Allocation and BERTopic to extract the recurring elements as well as K-means clustering to group folktales. \ Our paper tries to answer the question what are the similarities and differences between folktales, and what do they say about culture. \ Here we show that the common trends between folktales are family, food, traditional gender roles, mythological figures, and animals. Also, folktales topics differ based on geographical location with folktales found in different regions having different animals and environment. \ We were not surprised to find that religious figures and animals are some of the common topics in all cultures. However, we were surprised that European and Asian folktales were often paired together.
\ Our results demonstrate the prevalence of certain elements in cultures across the world. \ We anticipate our work to be a resource to future research of folktales and an example of using natural language processing to analyze documents in specific domains. Furthermore, since we only analyzed the documents based on their topics, more work could be done in analyzing the structure, sentiment, and the characters of these folktales. 

\end{abstract}

\section{Introduction}

From Momotaro to Pandora's Box, from Chinese zodiac to Frog King, there are many interesting stories passed down from generation to generation orally all over the world, and we call these stories: folktales. Since these folktales have very strong connections with their local culture, some clear differences between stories from different regions can often be seen, but there may also be some inner connections between them. To better dig out and analyze the hidden connections within these folktales from across the world, we use natural language processing techniques like topic modeling and clustering to analyze these stories.

Previously, there has been many researches using machine learning technologies to analyze the text structure of folktales. In \citet{Laudun:13}'s paper, they applied Latent Dirichlet Allocation on three folklore studies journals that were over 125 years old to understand the changes and flow of topics and paradigms within a domain. In Meder's work, they tried to use machine learning to let the computer learn the structure of folktales from the Dutch folktale database and generate meta folktales finally \citep{Meder:16}. In Agnihotri's work the researches prove the effect of k-means to cluster texts and help find the similarity between stories, news and emails \citep{Agnihotri:14}.

In this paper, we present an analysis of folktales from different regions using topic modeling technologies like LDA and BERTopic and clustering technology like K-means. We try to answer the following questions inside these folktales:
\\1) What are the topics that are popular in all cultures? 
\\2) Are there specific topics that are mentioned more in some regions?
\section{Methodology}
Our task includes two parts: topic modeling and clustering. For topic modeling, we used both Latent Dirichlet Allocation (LDA) and BERTopic to extract reoccurring elements of folktales of different regions and categorize relevant elements into similar topic groups. For clustering, we utilized K-means to group the original texts and analyze their relationships.

\subsection{Data}
We wrote a Python program to scrape and retrieve 726 folktales from \url{www.worldoftales.com}, which collects folktales from Project Gutenberg and categorizes them into different regional groups. 

We noticed a large difference in the number of stories between some regions. Due to this limitation, we decided to limit the number of stories from each region to 250 while gathering as many as we could for less represented regions. In addition, we tried to balance the number of stories per country in each region to ensure a diverse range of representation when possible. 

Here is the breakdown of our data:
\textbf{Africa (86 folktales): }Nigeria (40), South Africa (38), Tanzania (8);
\textbf{Asia (250): }The Arabs (40), China (83), Philippines (40), Japan (67);
\textbf{Australia (31)};
\textbf{Europe (250)}: England (42), France (40), Germany (30), Ireland (38), Italy (32), Russia (40), Sweden (28);
\textbf{North America/Native America (75)};
and \textbf{South America (34)}.

Since our program precisely scraped the text of the folktales, we did not need to transform the data too much. We put each story into a row and constructed two additional columns indicating the stories’ region and country respectively.

\subsection{Latent Dirichlet Allocation (LDA)}
To start, we applied an LDA approach with the Mallet model for topic modeling. Mallet is a Java-based package for statistical natural language processing such as topic modeling and information extraction (\citealp{McCallum:02}).

Before we used Mallet, we conducted data preprocessing, including stopword removal, tokenization, lemmatization, n-grams implementation, and speech of tag selection. To reduce noise in the model, we also extended the stopword list to include high-frequency words deemed not to have meaningful content. 

The Mallet model requires a dictionary and a corpus for processing. Therefore, we converted our data into a dictionary that contains all the words and their numeric token IDs, and then constructed a corpus using the dictionary. 

Unlike BERTopic, which automatically generates the optimal number of topics, LDA Mallet requires users to indicate the number of topics they want to extract in advance. To find the optimal number of topics for each region’s folktales, we computed the coherence score for different topic numbers and selected the one that had a relatively good score. Therefore, the number of topics generated for different regions may vary. 

The model returns only clustered terms for the set number of topics, along with the terms’ probabilities for each topic. This requires users to analyze the returned terms and label the topics manually. During the LDA topic modeling process, we analyzed all the topics, labeled them, and visualized the results.

\subsection{BERTopic}

BERTopic is a deep learning based topic modeling method which was first developed by Maarten Grootendorst \citeyearpar{https://doi.org/10.48550/arxiv.2203.05794}. It leverages BERT embeddings and c-TF-IDF to create dense clusters allowing for topics that are easy to interpret while at the same time keeping important words in the topic descriptions. It overcomes the limitations of previous methods including LDA, which disregard the semantic relationships among words, and thus should have a better performance doing topic modeling. 

In general, it takes three steps for BERTopic to generate topic representations. Firstly a pre-trained language model will convert each document to its embedding representation. Then, the algorithms reduce the dimensionality of the resulting embeddings to optimize the clustering process. Finally, topic representations are extracted from the clusters of documents using a custom class-based variation of TF-IDF \citep{https://doi.org/10.48550/arxiv.2203.05794}.

Here in order to feed our texts into the BERTopic network, we first split our stories into separate sentences. Since BERTopic uses document embeddings, there is typically no need to preprocess the data since all parts of the document are important to understand the general topic of the given document. Only when we provide documents containing lots of noise like HTML-tags we should do some cleaning of the original document before \citep{Bertopic:22}. We then feed in these sentences to the network as a group based on their original regions. BERTopic network will then try to cluster these sentences together and return clusters with different topics giving out some most represented key words under each of these topics. We then summarized the topics based on the given key words just like the method used in summarizing the topics in LDA.

\subsection{K-means Clustering}

K-means Clustering is an unsupervised learning algorithm and vector quantization method originally proposed by Stuart Lloyd \citeyearpar{Lloyd:82}. It separates observations into 'k' groups called clusters. Each cluster is formed around a centroid which is the mean of a cluster. These clusters are formed by finding the set of clusters that minimize the variance of data points within clusters. It is used to find patterns in data that indicate the presence of different groups in the data. It is able to group things based off of their similarities according to a set of features. 

The algorithm works by first randomly selecting 'k' centroids for potential clusters. Then it puts all the data points into the cluster which has the centroid closest to them. New centroids are then calculated by taking the average of all data points within a cluster and then this process is repeated for a selected 'i' iterations or until the centroids stop changing values. We selected the 'k' to be 10 and 'i' to be 10000. 

To generate the set of features for the clustering, a term frequency-inverse document frequency (tf-idf) matrix was generated from the raw texts of the folktales. This matrix determines the relevancy of words in a given folktale by combining the terms' frequencies, and the inverse document frequency which measures the rarity of a word across documents. Tf-idf was originally developed by Luhn \citeyearpar{luhn1957statistical} and Jones  \citeyearpar{jones1972statistical}. Principal Component Analysis (PCA), which was originally proposed by Hotelling, is then performed on the matrix \citep{Hotelling:33}. 
PCA transforms this matrix to a lower dimensional space by using an orthogonal transformation on it. This is to make the data more interpretable, i.e. visible in a 2d scatter plot, while not losing a significant amount of information from the data. K-means clustering is then performed on the PCA transformation to form clusters of folktales. This process re-run five times to confirm results.

\section{Results}
\subsection{Topic Modeling}

Both LDA Mallet and BERTopic were able to extract recurring elements of the folktales and categorize relevant elements into the same topic groups. A major difference between the two methods is that LDA sometimes returns clustered terms that seem unrelated and thus requires further interpretation. To understand and label some topics, we had to analyze the original folktales that contain those terms. Compared with LDA, the clustered terms that BERTopic returns are more precise and closely related to each other. However, since data preprocessing is not needed in the BERTopic method, there are a lot of stopwords that create a lot of noise in the results.

Therefore, to fully utilize the outputs of the two methods, we looked at the results of both methods together and considered the shared topics as the dominant topics for folktales in each region. 

\begin{figure}[htbp]
\centering
\includegraphics[width = 7cm]{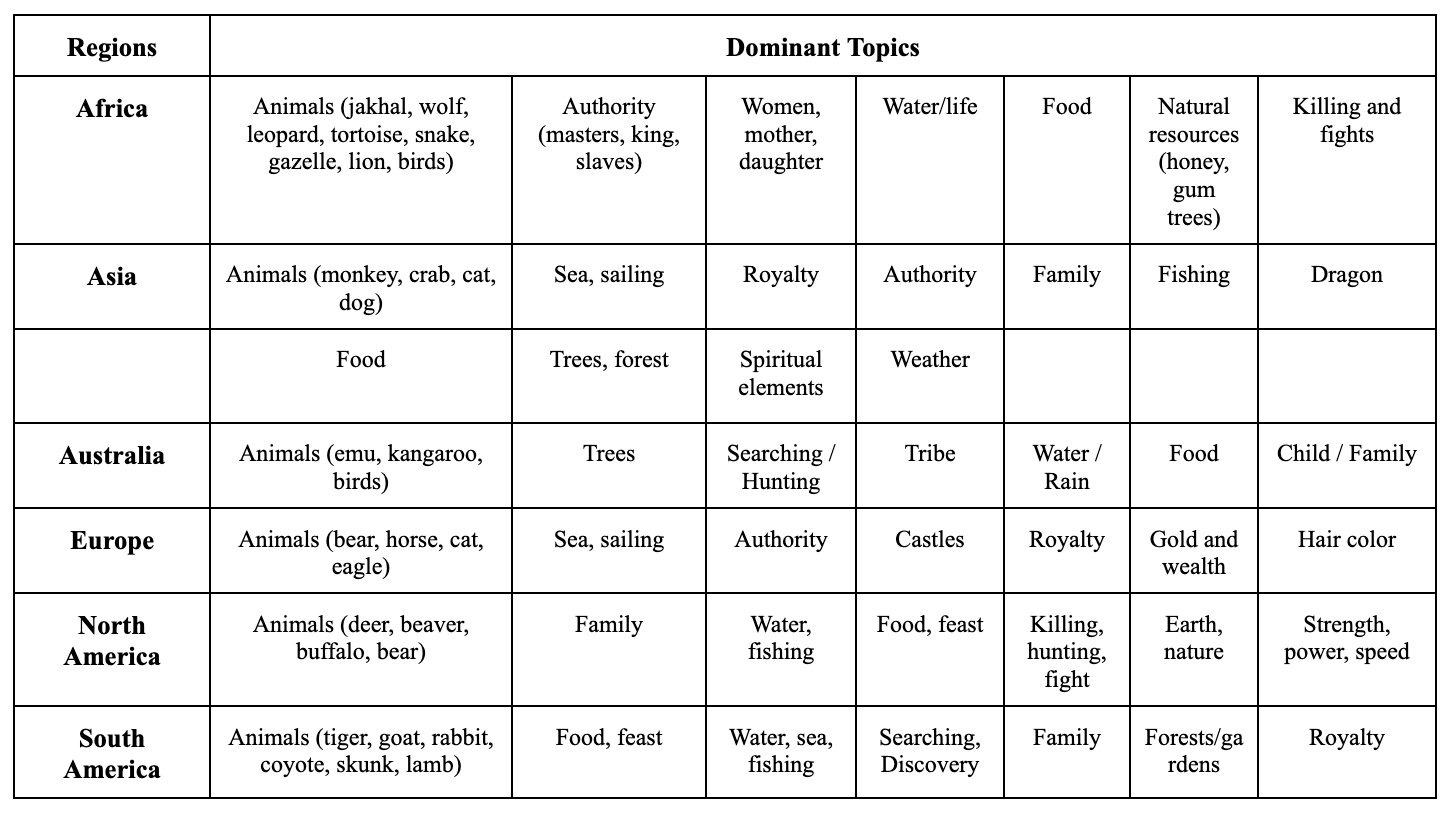}
\caption{\label{citation-guide}
Dominant topics of folktales in different regions. The results were summarized based on the outputs of both LDA and BERTopic methods. 
}
\end{figure}

As shown in Figure 1, family, food, animal, and water-related themes are major topics in folktales of all regions. Additionally, all regions have mythological and religious figures. Authority is a popular topic in African, Asian, and European folktales, while searching and hunting are mentioned a lot in Australia, North America, and South America.

Although all regions share some common topics, the specific content can be very different. For example, the particular animals in different regions’ folktales are different. Wild animals such as jackals, gazelles, and leopards appear a lot in African folktales, while Australian folktales talk more about emus and kangaroos. Even though food is a common topic, in Asian folktales, rice is the major source of food, while corn is more common in other regions including Europe, North America, and South America. Folktales across all regions have water-related elements, but African folktales revolve around the relationship between water and life, while Asia, Europe, South America focus on fishing, sailing, and sea. 

Other major differences are reflected on the region-specific elements. For instance, European folktales have more castles while Asian stories feature temples more often. Dragon is a distinctive feature in Asian stories, while hair colors such as "blondine" and "violette" are distinct elements in European folktales, and speed and strength are more often mentioned in North American folktales.

\subsection{Clustering}
After forming the clusters, we visualized them in four ways. The first way, in Figure 2(a), was a bar graph showing the amount of folktales in each cluster and the regions they come from. The second way, in Figure 2(b), was a bar graph that better visualizes the region breakdown within each cluster. The third way, in Figure 2(c), was a scatter plot of the clustered data. This shows and highlights the distances between different clusters. Lastly, a \href{https://drive.google.com/file/d/1lU-ljV62qgLH4uXiuVzLrEy3o7Ky1S9k/view}{hierarchical dendrogram} was created which shows an alternative way of viewing the distances between clusters.

\begin{figure}[htbp]
\centering
\subfigure[1]{
\begin{minipage}[t]{0.25\textwidth}
\centering
\includegraphics[width = 4cm]{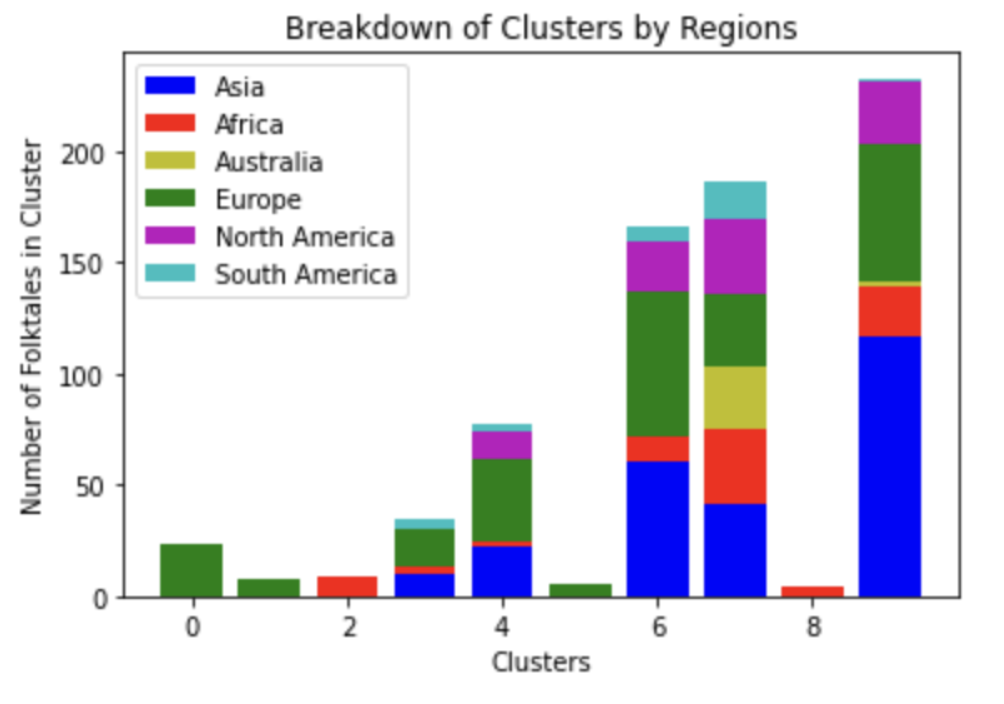}
%\caption{fig1}
\end{minipage}%
}%
\subfigure[2]{
\begin{minipage}[t]{0.25\textwidth}
\centering
\includegraphics[width = 4cm]{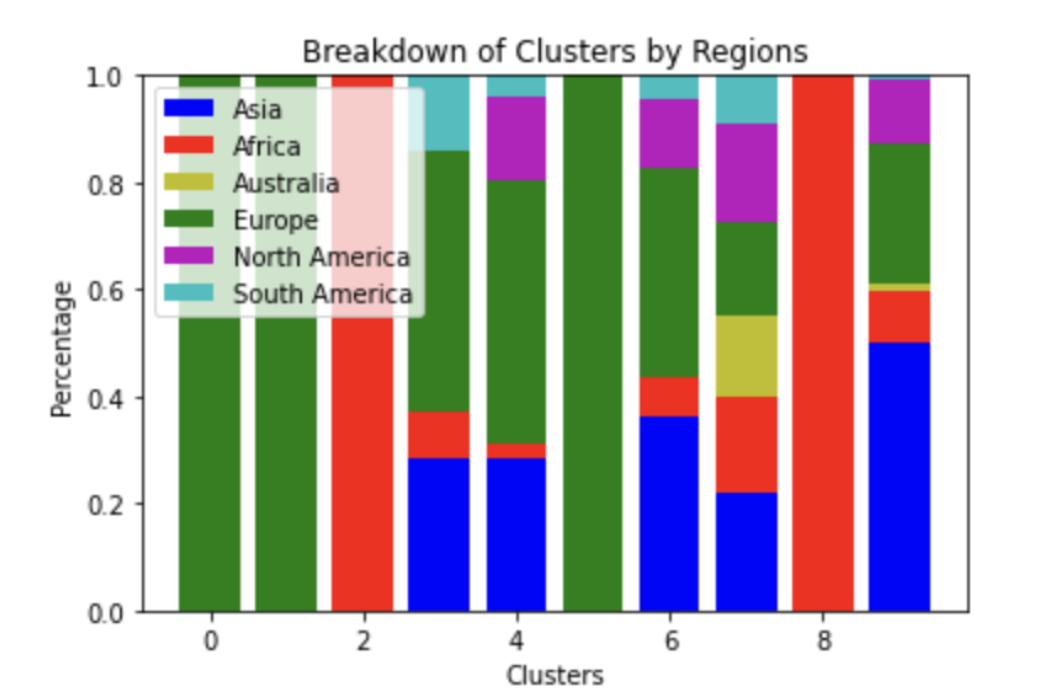}
%\caption{fig1}
\end{minipage}%
}%

\subfigure[3]{
\begin{minipage}[t]{0.4\textwidth}
\centering
\includegraphics[width = 4cm]{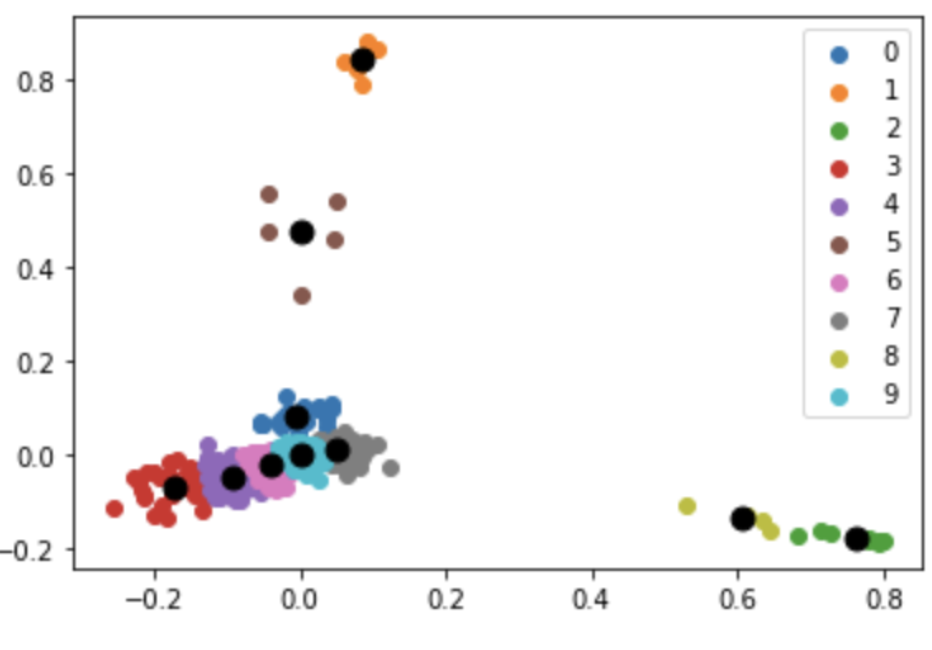}
%\caption{fig1}
\end{minipage}%
}%
\caption{\label{citation-guide}
Visualization of clustered folktales
}
\end{figure}

\section{Discussion}

The results of our topic modeling show common topics like princesses, daughters, and wives as well as young and old men and kings between all the regions. This also includes topics that emphasize traditional gender roles where men and women take on dominant and submissive roles respectively. These results could support research in gender studies since they reflect the prevalence of gender roles in culture.

One of the most common topics between regions was mythological figures which demonstrates the shared importance of mythology, i.e., some form of storytelling passed down from generation to generation, within all cultures. The breakdown of what mythological figures are prevalent in what regions could be used to support research in mythology.

The presence of different types of animals and geographical features like forests and plains demonstrate the impact of geography and environment on the cultural values of regions. For instance, jackals are a common topic in African folktales and jackals are found in Africa. This supports the idea that location and environment have an impact on cultural values. It also shows what are the key geographical and environmental influences on a region's culture. This could potentially be used to support research in cultural geography. 

From our clustering it appears that across the five runs, the Europe and Africa folktales are being consistently put into their own far off clusters. Europe being separate from the group is expected since it is our only Western region from our list of regions. The more peculiar one is Africa which appears the furthest out from all the rest. This is unexpected behavior since it does not appear with other non-Western nations. This complicates the idea of a Western vs. Non-Western split within the world.

In terms of ethical concerns, the topics themselves were manually created by us and that presents an opportunity for bias. Also, the data set was originally created by a source we do not have the methodology for who could have introduced their own bias in a number of ways. Also, the results of this study can be used to create products that target regions by the values revealed through these findings. This brings up the ethical concern of whether these topics should be used to create products that target the values of particular regions/cultures.

\section{Conclusions}
Using LDA and BERTopic, our results demonstrated the prevalence of certain values, such as family, food, mythological figures, animals, and traditional gender roles in folktales across different regions. They also showed the effect that geography has on culture. From K-means clustering, we discovered that European and African folktales often formed their own distinct clusters. When they did appear with other regions, it was often paired with Asian folktales. These common and differing topics found are reflective of the values of the cultures around the world. Since we only collected folktales from specific countries to represent a region and the number of folktales collected in each region are not equal, possible bias caused by our original data may exist. In the future, more comprehensive data sets should be used and more aspects of the stories, such as structure, sentiment, and characters should be explored.

\bibliography{anthology,custom}
\bibliographystyle{acl_natbib}

\end{document}